%% file: 0_main.tex
\pdfoutput=1
\documentclass[11pt]{article}

\usepackage[final]{acl}

\usepackage{times}
\usepackage{latexsym}

\usepackage[T1]{fontenc}
\usepackage[utf8]{inputenc}

\usepackage{microtype}

\usepackage{inconsolata}

\usepackage{graphicx}

\usepackage{multirow}
\usepackage{multicol}
\usepackage{booktabs} 
\usepackage{adjustbox}
\usepackage{amsmath}
\usepackage[T1]{fontenc}
\usepackage{cleveref}
\usepackage[ruled,vlined]{algorithm2e}
\usepackage{amsmath, amssymb}
\usepackage[normalem]{ulem} 
\usepackage{enumitem}
\usepackage[table]{xcolor}  % 启用颜色支持
\usepackage{url}

\newcommand{\ourmethodshort}{NeuroAda}
\title{NeuroAda: Activating Each Neuron's Potential for \\Parameter-Efficient Fine-Tuning}

\author{
  Zhi Zhang\thanks{Equal contribution.} \\
  ILLC, University of Amsterdam \\
  \texttt{zzhang2626@gmail.com} \\\And
  Yixian Shen\footnotemark[1] \\
  PCS, University of Amsterdam \\
  \texttt{y.shen@uva.nl} \\ \AND
  Congfeng Cao \\
  ILLC, University of Amsterdam \\
  \texttt{c.cao@uva.nl} \And
  Ekaterina Shutova \\
  ILLC, University of Amsterdam \\
  \texttt{e.shutova@uva.nl} \\
}

\begin{document}
\maketitle

\date{
\textsuperscript{*}equall contribution
}

\begin{abstract}
Existing parameter-efficient fine-tuning (PEFT) methods primarily fall into two categories: addition-based and selective in-situ adaptation. The former, such as LoRA, introduce additional modules to adapt the model to downstream tasks, offering strong memory efficiency. However, their representational capacity is often limited, making them less suitable for fine-grained adaptation. In contrast, the latter directly fine-tunes a carefully chosen subset of the original model parameters, allowing for more precise and effective adaptation, but at the cost of significantly increased memory consumption.
To reconcile this trade-off, we propose \ourmethodshort{}, a novel PEFT method that enables fine-grained model finetuning while maintaining high memory efficiency. Our approach first identifies important parameters (i.e., connections within the network) as in selective adaptation, and then introduces bypass connections for these selected parameters. During finetuning, only the bypass connections are updated, leaving the original model parameters frozen.
Empirical results on \textbf{23+} tasks spanning both natural language generation and understanding demonstrate that \ourmethodshort{} achieves state-of-the-art performance with as little as $\leq \textbf{0.02}\%$ trainable parameters, while reducing CUDA memory usage by up to \textbf{60\%}.
We release our code here: \url{https://github.com/FightingFighting/NeuroAda.git}.

\end{abstract}

\input{1_introduction}

\input{2_relatedwork}

\input{3_method}

\input{4_experiment}

\input{5_conclusion}

% \clearpage
\input{7_limitation}
\bibliography{emnlp,anthology}

\clearpage
\appendix

\input{6_appendix}

% This is an appendix.

\end{document}

%% file: 1_introduction.tex
\section{Introduction}
\label{sec:intro}

Large language models (LLMs) demonstrate remarkable generalization capabilities on various NLP tasks~\cite{dong-etal-2023-open,li-etal-2025-coir}, yet achieving optimal performance on downstream tasks often still requires fine-tuning. 
As model sizes grow, full-parameter fine-tuning becomes increasingly impractical due to substantial computational and memory demands. 
For example, fine-tuning LLaMA~2-13B without CPU offloading requires 26~GB for trainable parameters in FP16, 52~GB for Adam optimizer states (two FP32 moments per parameter), 26~GB for gradients, and an additional 2--4~GB for activations depending on batch size and sequence length. This results in a memory footprint of approximately 106--108~GB in total, far exceeding the capacity of commodity GPUs and necessitating premium hardware (e.g., A100 80G). 
This highlights the pressing need for more efficient and scalable adaptation strategies.

\begin{figure}
    \centering
    \includegraphics[width=1\linewidth]{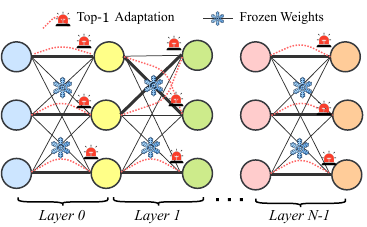}
    \caption{Overview of \ourmethodshort{}. For each neuron, top-$1$ weights are adapted, while the rest remain frozen. Bold dark indicates selected pretrained weights; red dashed edges represent newly introduced trainable parameters.
}
    \label{fig:mag}
\end{figure}

% (blue solid lines)
% (blue solid lines)
A growing body of work on parameter-efficient fine-tuning (PEFT) addresses the computational and memory overhead of full-model adaptation by introducing a set of trainable parameters while keeping the backbone frozen. 
One major class of these methods is known as \textit{addition-based adaptation}, which augments the pretrained model with additional modules designed to inject task-specific flexibility. 
These additions vary in form and location, including adapter layers inserted into projection blocks~\cite{pfeiffer2020adapterfusion, sung2022vl},
nonlinear activation reparameterizations~\cite{zhang2021tip}, prompt tuning applied to input embeddings~\cite{lin2020exploring, liao2023make, liao2023parameter}, 
latent representation perturbations~\cite{wu2024advancing}, 
and low-rank matrix decompositions applied directly to weight spaces, such as in LoRA~\cite{hu2022lora} and its variants~\cite{zhang2303adaptive, kopiczko2023vera}. 
These methods typically offer improved memory efficiency by restricting gradient computation and optimizer state updates to the newly introduced modules, as opposed to the entire model in full fine-tuning. However, their scalability is constrained: as model size increases, the limited representational capacity of the added modules often leads to diminishing returns~\cite{he_sparse_2024}.

\begin{figure}
    \centering
    \includegraphics[width=\linewidth]{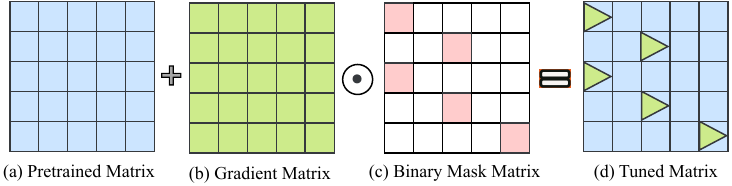}
    \caption{
Mask-based sparse tuning employs a binary mask matrix to suppress gradient updates for unselected parameters. However, this approach incurs significant memory overhead, as gradients for the entire original parameter matrix must still be computed and retained by the optimizer.
}
\vspace{-10pt}
\label{fig:sparsetuning}
\end{figure}

Another prominent line of research is \textit{selective in-situ adaptation}, which fine-tunes a carefully selected subset of a pretrained model's original parameters, without introducing any additional parameters, modules, or layers. Structure-based approaches, such as BitFit~\cite{ben_zaken_bitfit_2022} and Partial-$k$~\cite{jia_visual_2022} updating only the bias terms and the last $k$ layers, respectively, exemplify early efforts in this direction. 
More recently, fine-grained and unstructured parameter selection methods have attracted increasing attention. These approaches aim to identify task-relevant parameters at a more granular level, such as GPS~\cite{zhang2024gradient} and SPT~\cite{he_sensitivity-aware_2023}, which demonstrate strong performance on vision tasks by selectively fine-tuning subsets of parameters that are most critical for the target task. Compared to structured approaches and \textit{addition-based adaptation} methods, these unstructured strategies offer greater flexibility in parameter selection, enabling more precise and targeted model adaptation, and thereby substantially improving downstream performance~\cite{shen2024expanding,fu2023effectiveness}. However, this sparse tuning paradigm leads to higher memory usage due to mask-based implementations. As shown in 
\Cref{fig:sparsetuning}, although only a small portion of the parameters is selected for updating, memory consumption remains comparable to that of full fine-tuning.
This limitation becomes particularly problematic with the increase of model size~\cite {zhai2022scaling,shen2024expanding,dong2025mmdocrag}.

These limitations motivate us to explore whether a unified approach can be designed that achieves the fine-grained parameter tuning characteristic of selective in-situ adaptation, while retaining the memory efficiency advantages of addition-based methods.
To this end, we propose \textbf{\ourmethodshort}, an additive, overlay-style adaptation method that utilizes a carefully designed approach to introduce new parameters to enable fine-grained adjustments while maintaining memory efficiency. Specifically, 
% that activates each neuron's potential through featherlight updates.
as shown in Figure~\ref{fig:mag}, \ourmethodshort{} first selects the top-$k$ highest-magnitude input connections (weights/parameters) for each neuron in the network prior to finetuning and then, for each selected parameter, a bypass connection (initialized to zero) is introduced. During finetuning, only the newly introduced parameters are fine-tuned, while the original parameters remain frozen. 
This approach inherits both the performance benefits of ~\textit{selective in-situ adaptation} and the memory efficiency of \textit{addition-based methods}.  Crucially, for each neuron, at least one of its input connections is selected for update, ensuring that all neurons have the potential to modify their activation states and thus change the state of the entire network for effective adaptation. 
Consequently, \ourmethodshort{} presents a scalable and practical solution for large LLMs.

\ourmethodshort{} offers four key advantages that make it both practical and effective in real-world scenarios:  
 (1)~\textbf{Highly efficient computation:}  
\ourmethodshort{} eliminates the need of the mask for sparse finetuning, which typically requires full gradient computation. 
(2)~\textbf{Highly efficient GPU memory usage:} Only the newly added parameters are updated, significantly lowering memory usage by avoiding optimizer state tracking for the full model.
(3)~\textbf{Task-agnostic and generalizable:}  
Parameter selection is based on weight magnitudes from the pretrained model, enabling consistent selection across tasks, making the method broadly applicable and easy to deploy.
(4)~\textbf{Fine-grained, neuron-level adaptation:} 
\ourmethodshort{} ensures every neuron has the potential to change the activation state of each neuron during finetuning, maximizing the representational expressiveness of individual neurons. 
Empirically, \ourmethodshort{} achieves state-of-the-art performance on 23+ tasks compared with other PEFT methods, including both natural language generation and understanding, highlighting its practical effectiveness.

%% file: 2_relatedwork.tex
\section{Related Work}

\paragraph{Addition-based Adaptation.} Includes adapter-based methods~\cite{he2021towards, pfeiffer2020adapterfusion, lin2020exploring, liao2023make, liao2023parameter} and low-rank reparameterization techniques such as LoRA~\cite{hu2022lora} and its variants AdaLoRA~\cite{zhang2303adaptive}, VeRA~\cite{kopiczko2023vera}, QLoRA~\cite{dettmers2024qlora}, and DoRA~\cite{liu2024dora}, which introduce trainable low-rank matrices into projection layers. While LoRA avoids inference-time overhead by merging updates into base weights, it often suffers from scalability issues and diminishing returns when applied to large models or complex tasks~\cite{liu2024dora}. A parallel direction modifies hidden states instead of weights: activation steering~\cite{liu2023context, li2023inference}, concept erasure~\cite{belrose2023leace, avitan2024changed, singh2024mimic}, and block-level editing~\cite{wu2024advancing} offer instance-specific control but require task-specific adaptation.

\paragraph{Selective In-Situ Adaptation.}  
This class of work fine-tunes a subset of the model’s original parameters without introducing any additional modules or weights, often achieving strong performance with minimal architectural changes. 
However, its practical memory and compute benefits frequently fall short in large-scale settings. 
Methods such as SIFT~\cite{song2023sparse}, SHiRA~\cite{bhardwaj2024sparse}, SpIEL~\cite{ansell2024scaling} and work from~\citep{zhang2024proactive} enforce sparsity constraints, yet still require full backward passes to compute gradients for the entire weight space.
More targeted approaches, including SMT~\cite{he_sparse_2024} and GPS~\cite{zhang2024gradient}, improve efficiency by selecting submatrices or top-$k$ gradients per neuron, but rely on gradient-based warm-up and dynamic masking. 
These mechanisms introduce additional overhead from binary mask storage, dense optimizer states, and full-gradient tracking, making them difficult to scale to large language models. In contrast, our method,~\ourmethodshort{}, inherits the advantages of both paradigms by avoiding gradient-based selection and selecting the top-$k$ weights per neuron.

%% file: 3_method.tex
\section{Methodology}
\label{sec:method}

\begin{figure}
    \centering
   \includegraphics[width=\linewidth]{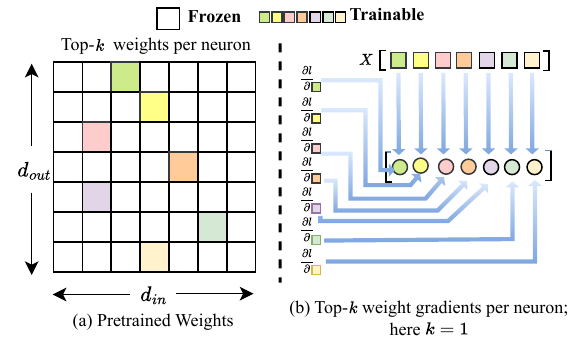}
    \caption{Neuron-wise Top-$k$ Weight Selection and Gradient Computation.
    (a) Pretrained weight matrix of size $d_{out} \times d_{in}$, where for each neuron (row), only the top-$k$ weights(i.e., highest-magnitude) are selected for adaptation (colored), and the rest remain frozen (white). 
    (b) Corresponding gradient matrix restricted to the top-$k$ weights per neuron (here $k = 1$), showing gradients only for trainable entries.
    This strategy enables fine-grained, neuron-level adaptation while preserving most of the pretrained model, effectively activating each neuron's potential through less-invasive tuning.}
    \label{fig:topk_gradients}
\end{figure}

% \textcolor{red}{the new introduced parameters are initialized to zero}

In this section, we first present the necessary preliminaries, and then introduce \ourmethodshort, 
a new adaptation framework that activates each neuron's potential by selectively updating a small subset of its weights. Specifically, we freeze all pretrained model weights and introduce sparse, additive overlay-style adaptation method in which top-$k$ bypasses of input connections (weights/parameters) are introduced into each neuron in the neural network for adaptation. This neuron-wise adaption preserves the original parameters intact while enabling targeted learning signals at fine granularity. 
% Specifically, we freeze all pretrained model weights and introduce sparse, additive updates-overlay-style-to the top-$k$ elements of each neuron.
% This neuron-wise delta-based adaptation preserves the original parameters intact while enabling targeted learning signals at fine granularity. 
As illustrated in Figure~\ref{fig:topk_gradients}, \ourmethodshort{} ensures that every neuron participates in adaptation, supporting both efficiency and generalization. 
During inference, the small number of learned deltas can be merged into the base weights, resulting in no additional overhead at runtime.

\subsection{Preliminaries}
\label{sec:prelim}

Let $\mathcal{M}_{\boldsymbol{\Phi}}$ be an $L$-layer pretrained language model with parameters
$\boldsymbol{\Phi}=\{\mathbf{W}^{(\ell)},\mathbf{b}^{(\ell)}\}_{\ell=1}^{L}$.
For any linear sub-layer we write
$\mathbf{h}_{\text{out}}=\mathbf{W}\mathbf{h}_{\text{in}}+\mathbf{b}$, where
$\mathbf{W}\!\in\!\mathbb{R}^{d_\text{out}\times d_\text{in}}$ and each
\emph{row} of~$\mathbf{W}$ corresponds to a \textbf{neuron}.
During standard fine-tuning all entries of~$\mathbf{W}$ are updated, yielding
heavy computational and memory costs (§\ref{sec:intro}).

\paragraph{Sparse additive updates.}
\ourmethodshort{} freezes $\boldsymbol{\Phi}$ and introduces a
\emph{delta‐parameter tensor} $\boldsymbol{\Delta}$ with the \emph{same
shape} as~$\boldsymbol{\Phi}$ but \emph{sparsity constrained}:

\begin{small}
\begin{equation}
\boldsymbol{\Phi}' \;=\; \boldsymbol{\Phi}\;+\;\boldsymbol{\Delta},
\qquad
\|\boldsymbol{\Delta}\|_0 \ll \|\boldsymbol{\Phi}\|_0,
\end{equation}
\end{small}
where $\|\cdot\|_0$ counts non-zero elements.  
Only $\boldsymbol{\Delta}$ is trainable; the base model remains intact, so
$\boldsymbol{\Delta}$ can be \emph{merged in-place} after training, incurring
zero inference overhead.

\subsection{Top-\texorpdfstring{$k$}{k} selection}
\label{sec:topk}
A core design goal is that \emph{every neuron receives at least a small learning signal}.  
For each neuron—that is, for each row
$\mathbf{w}\in\mathbb{R}^{d_\text{in}}$ of a weight matrix,
we identify the indices of its $k$ largest-magnitude components:

\begin{small}

\begin{equation}
\mathcal{I}(\mathbf{w}) \;=\;
\operatorname*{arg\,top\;k}_{j\in \{1,\dots,d_\text{in}\}}
\bigl|\mathbf{w}_{j}\bigr|.
\end{equation}
    
\end{small}
We then allocate trainable delta weights only at these positions:

\begin{small}
\begin{equation}
\bigl[\boldsymbol{\Delta}\bigr]_{i,j} \;=\;
\begin{cases}
\theta_{i,j} & \text{if } j\in\mathcal{I}(\mathbf{w}_{i}) \\[4pt]
0            & \text{otherwise},
\end{cases}
\label{eq:sparse-delta}
\end{equation}
\end{small}
where $\theta_{i,j}$ is a trainable parameter defined only for $j \in \mathcal{I}(\mathbf{w}_i)$ and initialized to $0$. 
For all other positions, $\boldsymbol{\Delta}{i,j}$ is fixed to zero and excluded from both optimization and memory storage.
While \ourmethodshort{} uses weight magnitude for top-$k$ selection, the framework is flexible: task-guided criteria such as gradient magnitude or random ticketing can be substituted into $\mathcal{I}(\cdot)$. We employ magnitude due to its task-agnostic stability and the advantage of requiring no warm-up or additional computation.
This design choice is empirically validated in our ablation study, where magnitude-based selection achieves strong performance without relying on task-specific signals, as shown in \Cref{fig:different_add_selection_method}.

% \footnote{%
% Other task-guided criteria (e.g.\ gradient magnitude or random tickets) can be
% plugged into $\mathcal{I}(\cdot)$, but magnitude works reliably across
% tasks and needs \emph{no warm-up}.}

\begin{table}[t]
\centering
\small
\setlength{\tabcolsep}{6pt}
\renewcommand{\arraystretch}{1.1}
\caption{
Memory comparison per projection.
Mask-based sparse tuning methods require 1 bit per weight\protect\footnotemark.
\ourmethodshort{} with $k=1$ only stores one BF16 value (2 bytes) and one index (2 bytes) per row, totaling 4 bytes per neuron.
This yields over $100\times$ memory savings for a single linear layer.
}
\resizebox{1\linewidth}{!}{
\begin{tabular}{lcccc}
\toprule
\textbf{Model} & $d_{\text{model}}$ &
\textbf{Mask [MB]} &
\textbf{\ourmethodshort{}[MB]} &
\textbf{Saving Ratio} \\
\midrule
LLaMA-1 7B     & $4096$  & $\frac{4096^2}{8 \times 2^{20}} \approx 2.00$ & $\frac{4096 \times 4}{2^{20}} \approx 0.016$ & $\approx 125\times$ \\
LLaMA-2 7B     & $4096$  & $2.00$                                      & $0.016$                                     & $125\times$ \\
LLaMA-1 13B    & $5120$  & $\frac{5120^2}{8 \times 2^{20}} \approx 3.13$ & $\frac{5120 \times 4}{2^{20}} \approx 0.020$ & $\approx 156\times$ \\
LLaMA-2 13B    & $5120$  & $3.13$                                      & $0.020$                                     & $156\times$ \\
\bottomrule
\end{tabular}
}

\label{tab:memory_k1}
\end{table}

\footnotetext{While 1-bit-per-weight is a theoretical lower bound for binary mask storage, actual implementations in PyTorch and other frameworks use byte-addressable storage (e.g., \texttt{BoolTensor}), leading to significantly higher memory overhead.}

\paragraph{Mask-free implementation.}
Since the top-$k$ sparsity pattern is determined \emph{a priori}, Eq.~\eqref{eq:sparse-delta} can be implemented without maintaining a full binary mask over the weight matrix. 
Instead, we store a compact list of \emph{indices} and corresponding \emph{BF16 values}—only $k$ entries per row—eliminating the need for dense masking or indexing during training. 
This design leads to substantial memory savings and indexing efficiency. 
As shown in Table~\ref{tab:memory_k1}, for a single projection layer in LLaMA-2 13B, a 1-bit-per-weight binary mask requires over 3~MB of memory, while \ourmethodshort{} with $k{=}1$ uses only 0.02~MB—over \textbf{156$\times$} smaller.
These savings scale across layers and are especially beneficial for high-throughput training on limited-memory devices.

\subsection{Featherlight adaptation}
\label{sec:adapt}

During fine-tuning we optimize only $\{\theta_{i,j}\}$ while re-using the
forward path of the frozen backbone.  
For a linear layer the forward pass becomes

\begin{small}
\begin{equation}
\mathbf{h}_{\text{out}}
\;=\;
\underbrace{\mathbf{W}\mathbf{h}_{\text{in}}}_{\text{frozen}}
\;+\;
\underbrace{
\bigl(\mathbf{P}\!\odot\!\boldsymbol{\Theta}\bigr)\mathbf{h}_{\text{in}}
}_{\text{trainable $\;\boldsymbol{\Delta}$}},
\end{equation}
    
\end{small}
where $\mathbf{P}$ is an index matrix with zeros everywhere except
$\bigl[P\bigr]_{i,j}=1$ when $(i,j)\!\in\!\mathcal{I}(\mathbf{w}_{i})$,
$\odot$ denotes element-wise product, and $\boldsymbol{\Theta}$ is the dense
tensor of learnable $\theta_{i,j}$.\footnote{We implement this with fused
scatter-add so the additional multiply is executed only on the $k$ selected
positions; no dense mask is materialised.}

\paragraph{Lightweight backward pass and optimizer states.}
During back-propagation, \ourmethodshort{} updates only the $k$ selected coordinates per neuron. 
As a result, the dominant memory contributors in full-model training—BF16/FP32 gradients and the two FP32 moment estimates in the AdamW optimizer—are reduced proportionally by a factor of $\frac{k}{d_{\text{in}}}$. 
Because all delta parameters are stored directly in BF16 and no FP32 master weights are needed, the optimizer maintains only $2 \times k$ FP32 values per row instead of $2 \times d_{\text{in}}$.This yields a substantial memory reduction in the optimizer state maintained by AdamW. In standard dense fine-tuning, AdamW stores two FP32 moment estimates (first and second moments) for each trainable parameter, resulting in:

\vspace{-13pt}
\begin{small}
\begin{equation}
\text{AdamW Mem. (Masked):} \quad
2 \times d_{\text{out}} \times d_{\text{in}} \times 4 \text{(bytes)},
\end{equation}    
\end{small}
\vspace{-15pt}

where 4 bytes denotes the size of a 32-bit float. In contrast, \ourmethodshort{} updates only $k$ weights per row, so the optimizer state becomes:

\vspace{-13pt}
\begin{small}
\begin{equation}
\text{AdamW Mem. (\ourmethodshort):}  2 \times d_{\text{out}} \times k \times 4\text{(bytes)}.
\end{equation}
\end{small}
\vspace{-15pt}

This reduces memory usage by a factor of $\frac{d_{\text{in}}}{k}$ per linear layer. For example, with $d_{\text{in}} = 5120$ and $k = 1$, this corresponds to a $5120\times$ reduction in AdamW state memory.

%% file: 4_experiment.tex
\section{Neuron-wise Sparse Adaptation: Comparative Analysis}
\label{sec:Preliminary experiments}
In this section, we first compare the mask-based method, which applies binary masks to zero out the gradients of unselected (frozen) parameters (see \Cref{fig:sparsetuning}), with our \ourmethodshort{}, which introduces new trainable parameters to bypass the selected ones, rather than directly tuning them, for sparse fine-tuning. The comparison is conducted in terms of effectiveness, GPU memory usage, and training efficiency.
We then further investigate the effectiveness of the proposed method \ourmethodshort{}, which aims to ensure that all neurons in the network have the potential to update their activation states during fine-tuning. This is done by analyzing the proportion of neurons involved in fine-tuning and examining different parameter selection strategies for activating them.

\begin{figure}[t]
    \centering
    \includegraphics[trim=16 23 16 15, clip, width=0.5\linewidth]{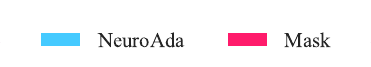}\\
    \includegraphics[trim=16 17 16 15, clip, width=0.48\linewidth]{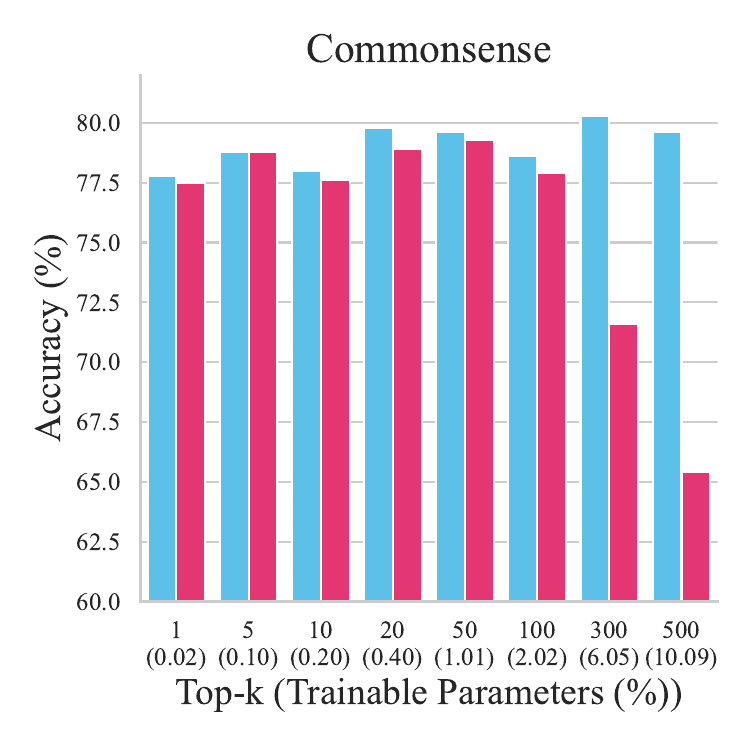}
    \includegraphics[trim=16 17 16 15, clip, width=0.48\linewidth]{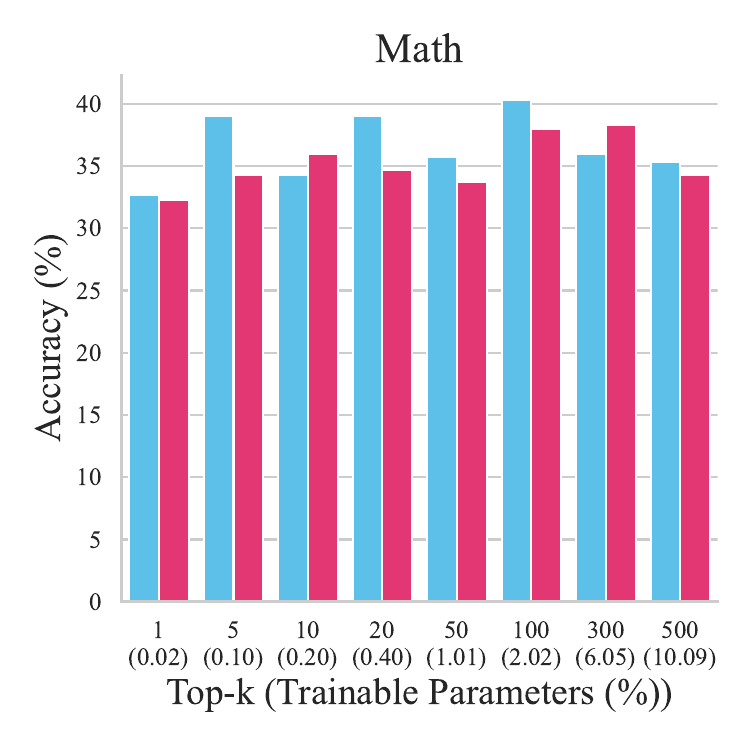}
    \caption{Performance comparison between our  \ourmethodshort{} and mask-based methods on LLaMA-7B. Top-$k$ means selecting top-$k$ input connections per neuron in the neural network.}
    \label{fig:mask_vs_add}
\end{figure}
\paragraph{Experiment setup} 
To ensure a fair comparison between our \ourmethodshort{} and mask-based methods, as well as across different parameter selection strategies, we conduct a hyperparameter search over the different learning rates for each experiment using the training set. The best-performing configuration is then selected based on validation set performance. This is necessary because PEFT methods are generally sensitive to the choice of learning rate~\cite{wu_reft_2024}. The hyperparameter search space is presented in \Cref{tab:llama_hyperparameters_different_trainable_num} in Appendix. The details of used datasets: \textsc{Commonsense15k} and \textsc{GSM8K} are provided in Appendix~\ref{sec:commondata}.

\paragraph{Question 1:}\textit{Can our method \ourmethodshort{} be a competitive or even superior alternative to the mask-based sparse tuning approach?} We address this by comparing their task performance, GPU memory usage, and training efficiency.

\paragraph{Performance} 
To comprehensively and fairly evaluate the effectiveness of the two methods, we compare them under the same proportion of trainable parameters, ranging from 0.02\% to 10\%, on the \textsc{Commonsense15k} and \textsc{GSM8K} tasks.
As shown in \Cref{fig:mask_vs_add}, our proposed method \ourmethodshort{} performs comparably to, or even better than, the mask-based method across most parameter budgets on both datasets.
In particular, the \ourmethodshort{} outperforms the mask-based method by approximately 9\% and 14\% in accuracy when using 6.05\% and 10.09\% of trainable parameters, respectively, on the commonsense reasoning task.

\begin{figure}[t]
    \centering
    \includegraphics[trim=0 23 0 15, clip, width=0.6\linewidth]{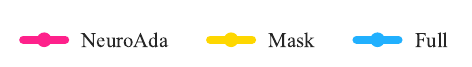} \\
    \includegraphics[trim=16 17 16 15, clip, width=0.48\linewidth]{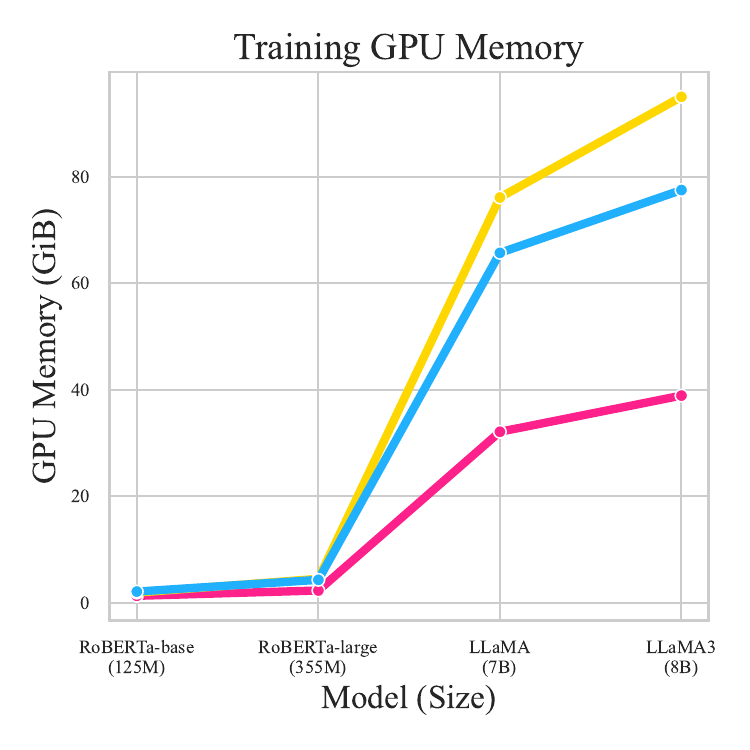}
    \includegraphics[trim=16 17 16 15, clip, width=0.48\linewidth]{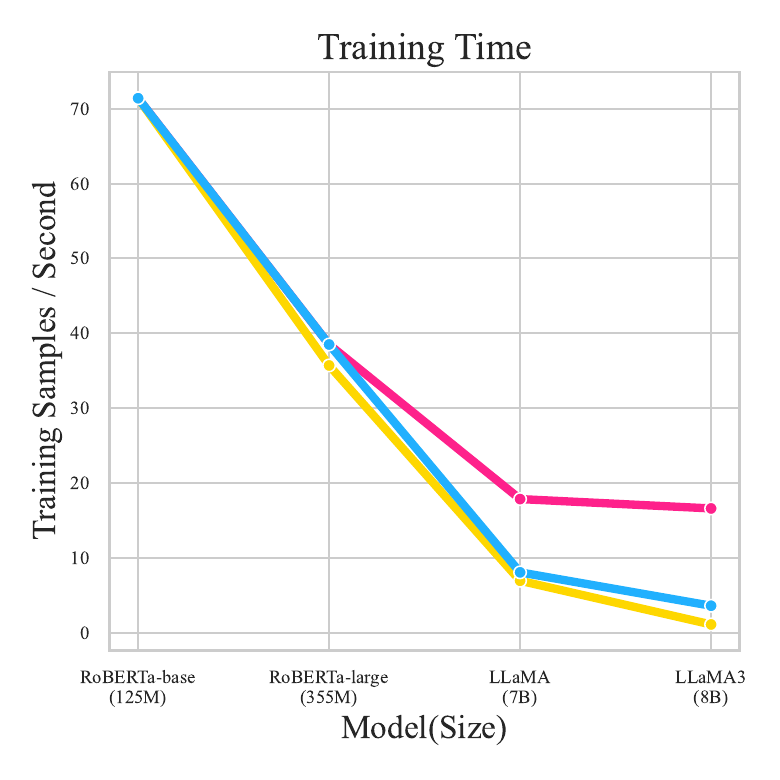}
    \caption{Training GPU memory and training efficiency on different models (RoBERTa-base, RoBERTa-large, LLaMA-7B, LLaMA3-8B) with \ourmethodshort{}, mask-based and full fine-tuning method.}
    \label{fig:memory_time_comparison}
\end{figure}

\paragraph{Training memory and time} 
We evaluate models of varying sizes, including RoBERTa-base, RoBERTa-large~\cite{liu2019roberta}, LLaMA-7B~\cite{touvron2023llama}, and LLaMA3-8B~\cite{vavekanandllama}. Specifically, we sample 500 examples from the MNLI task in the GLUE natural language understanding benchmark~\cite{wang2019glue}, and another 500 examples from the natural language reasoning task \textsc{GSM8K}. We use these samples to train the RoBERTa and LLaMA models, respectively.
All experiments are conducted on a single NVIDIA H100 GPU with a batch size of 2. We report the GPU memory usage and training time for each model.
\Cref{fig:memory_time_comparison} shows our proposed addition-based sparse training method \ourmethodshort{} consumes less GPU memory compared to the mask-based counterpart, especially as the model size increases. For example, the \ourmethodshort{} achieves up to 60\% memory savings on LLaMA3-8B.
In addition, the \ourmethodshort{} enables significantly faster training, particularly for larger models. It processes 16.6 samples per second, compared to only 1.1 samples per second with the mask-based method.

\paragraph{Question 2:}
\textit{How effective is the proposed method \ourmethodshort{} in enabling all neurons to update their activation states during fine-tuning for downstream task adaptation?}
To answer this question,  
we first investigate how different parameter selection strategies can be used to ensure that all neurons have the potential to update their activation states during fine-tuning.
We further analyze how task performance on \textsc{Commonsense15k} and \textsc{GSM8K} varies with the proportion of neurons allowed to update their activation states during training.

\begin{figure}[t]
    \centering
    \includegraphics[trim=16 17 5 15, clip, width=0.48\linewidth]{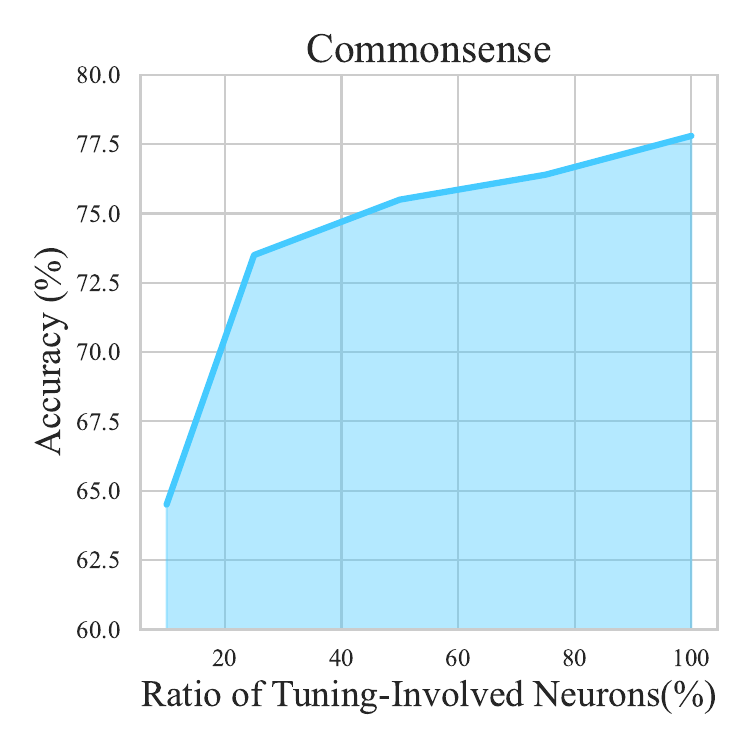}
    \includegraphics[trim=16 17 9 15, clip, width=0.48\linewidth]{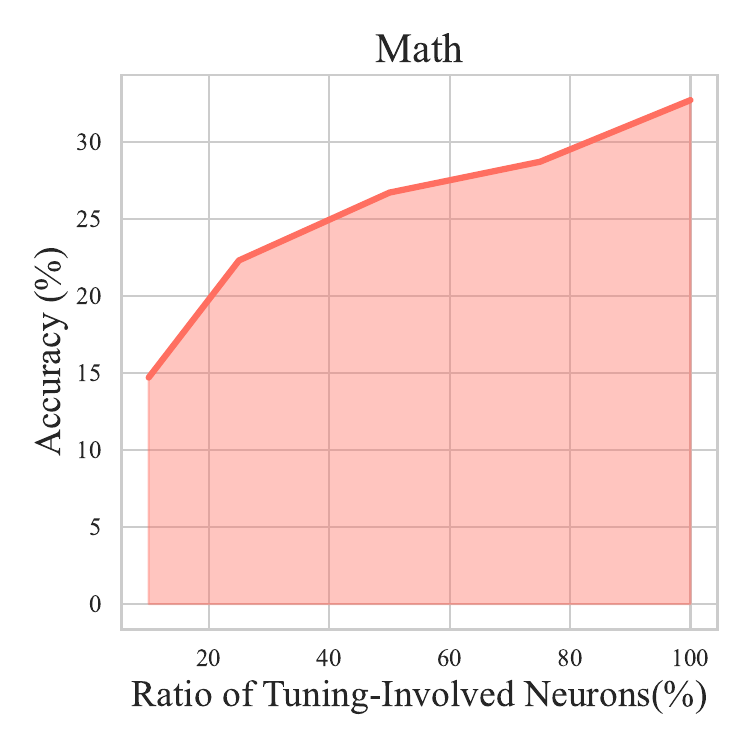}
    \caption{Comparison across different proportions of neurons involved in the fine-tuning process.}
    \label{fig:proportionOfInvolvedNeuron}
\end{figure}

\paragraph{Involved number of neurons.} 
Our proposed method selects the top-$k$ input connections for each neuron in the network, ensuring that at least one input connection per neuron is selected for tuning. This design enables all neurons to update their activation states during fine-tuning, allowing better adaptation to downstream tasks.
To demonstrate its effectiveness, we select parameters from various proportion of neurons per layer for tuning and evaluate the resulting performance on the \textsc{Commonsense15k} and \textsc{GSM8K} tasks.
As shown in \Cref{fig:proportionOfInvolvedNeuron}, increasing the number of neurons involved during training leads to consistent performance improvements on both tasks. This suggests that enabling all neurons to update their activation states is beneficial—and likely necessary—for effective downstream task adaptation.

\begin{figure}[t]
    \centering
    \includegraphics[trim=5 30 5 30, clip, width=0.9\linewidth]{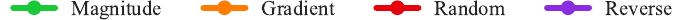}\\
    \includegraphics[trim=16 17 5 3, clip, width=0.48\linewidth]{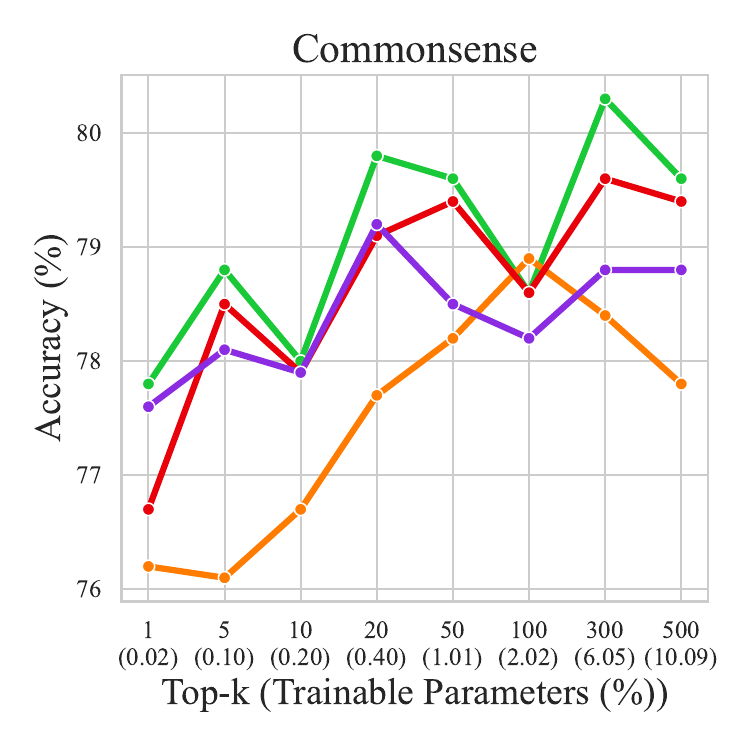}
    \includegraphics[trim=16 17 5 3, clip, width=0.48\linewidth]{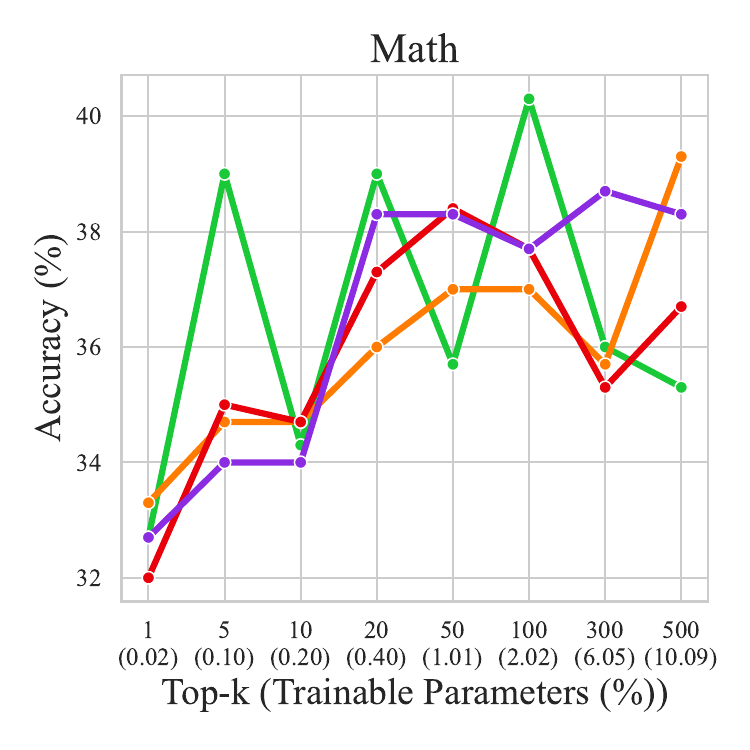}    
    \caption{Comparison of different parameter selection strategies for involving neurons for the fine-tuning process. 
    Among all input connections for each neuron in the network, Top-$k$ connection with highest magnitude (\textit{Magnitude}), highest gradient absolute value (\textit{Gradient}), lowest magnitude (\textit{Reverse}) are selected for training using addition-based method. \textit{Random} means randomly selecting Top-$k$ input connections per neuron.}
    \label{fig:different_add_selection_method}
\end{figure}

\paragraph{Different selection strategies} 
To explore the effectiveness of enabling all neurons in the network to update their activation states during training, we experiment with different parameter selection strategies for each neuron under different trainable parameter budget. Specifically, for each neuron, we select the top-$k$ input connections based on four criteria: highest weight magnitude, highest gradient absolute value, lowest weight magnitude, and random selection from all input connections. 
As shown in \Cref{fig:different_add_selection_method}, all selection methods yield comparable performance on both the \textsc{Commonsense15k} and \textsc{GSM8K} tasks across different trainable parameter budgets. The average accuracies across all budgets for all selection methods are closely aligned, ranging from 77.69\% to 79.24\% on \textsc{Commonsense15k}, and from 35.89\% to 36.54\% on \textsc{GSM8K}. 
These results again highlight the importance of involving all neurons in the adaptation process, regardless of the specific selection method used.
Additionally, across both tasks, the \textit{Magnitude} selection method achieves the highest win rate across different parameter budgets compared to the other strategies.
Therefore, we adopt the \textit{Magnitude} selection strategy as the default in \ourmethodshort{}.

\begin{table*}[!ht]
    \centering
    \caption{
    Performance comparison with existing PEFT methods on eight commonsense reasoning datasets across four models: LLaMA-7B/13B, LLaMA2-7B, and LLaMA3-8B. 
    $^*$Most baseline results are taken from \citet{hu_llm-adapters_2023}. $^\dagger$Results are from \citet{wu_reft_2024}, $^\ddagger$ results are taken from \citet{he_sparse_2024} and $^\star$results are from \citet{liu2024dora}, as they share the same experimental setting with \citet{hu_llm-adapters_2023}. For a fair comparison, our \ourmethodshort{} is also trained for 3 epochs to align with these baselines.
    $^\textbf{+}$When 3-epoch results are not available in the original paper, we re-trained the baselines using the official code and their reported best hyperparameters. All results for our method are averaged over three runs with different random seeds. Our method selects the top-20 and top-1 input connections per neuron for high-budget and low-budget parameter groups, respectively.
    }
    \label{tab:commonsense_result}
    \resizebox{1\linewidth}{!}{
    \begin{tabular}{clcccccccccl}
        \toprule
        \hline 
        \rowcolor{cyan!30}
        \textbf{Model} & \textbf{PEFT} & \textbf{Params (\%)} & \multicolumn{9}{c}{\textbf{Accuracy} ($\uparrow$)} \\
        % \multirow{2}{*}{\textbf{Model}} & \multirow{2}{*}{\textbf{PEFT}} & \multirow{2}{*}{\textbf{Params}} & \multicolumn{8}{c}{\textbf{Accuracy} ($\uparrow$)} \\
        \midrule \rowcolor{yellow!30}
        \multicolumn{12}{l}{\hspace{12cm}\textbf{Commonsense Reasoning}} \\
        \cline{4-12}  \rowcolor{yellow!30}
        & & & \textbf{BoolQ} & \textbf{PIQA} & \textbf{SIQA} & \textbf{HellaS.} & \textbf{WinoG.} & \textbf{ARC-e} & \textbf{ARC-c} & \textbf{OBQA} & \textbf{Avg.} \\
        \midrule
        ChatGPT$^*$& $\quad-$ & $-$ & 73.1 & 85.4 & 68.5 & 78.5 & 66.1 & 89.8 & 79.9 & 74.8 & 77.0 \\ \cmidrule{1-12}
        & Series$^*$& 1.953\% & 63.0 & 79.2 & 76.3 & 67.9 & 75.7 & 74.5 & 57.1 & 72.4 & 70.8 \\
        & Parallel$^*$& 3.542\% & 67.9 & 76.4 & 78.8 & 69.8 & 78.9 & 73.7 & 57.3 & 75.2 & 72.3 \\
        & LoRA$^*$& 0.826\% & 68.9 & 80.7 & 77.4 & 78.1 & 78.8 & 77.8 & 61.3 & 74.8 & 74.7 \\
        & DoRA$_{\text{half}}$$^\star$ & 0.427\% & {70.0} & 82.6 & 79.7 & 83.2 & 80.6 & 80.6 & 65.4 & 77.6 & 77.5 \\
        & DoRA$^\star$& 0.838\% & 68.5 & 82.9 & 79.6 & 84.8 & 80.8 & 81.4 & 65.8 & 81.0 & 78.1 \\ 
        LLaMA& SMT$^\ddagger$& 0.840\% &68.7 &81.7 &78.3 &91.6&78.8&84.1&67.7&77.4&78.7 \\ 
        \cmidrule{2-12}
        (7B)& \textbf{\ourmethodshort{}} & \textbf{0.404\%} &  \textbf{73.1}&\textbf{85.4} &\textbf{80.9} &\textbf{94.3}  &\textbf{84.3}  &\textbf{87.8}  &\textbf{71.7}  &\textbf{84.2} &\colorbox{cyan!30}{\textbf{82.7}} \\ \cmidrule{2-12}
        & PrefT$^*$& 0.039\% & 64.3 & 76.8 & 73.9 & 42.1 & 72.1 & 72.9 & 54.0 & 60.6 & 64.6 \\
        & DiReFT$^\textbf{+}$& 0.031\% & 66.1   &82.5   &78.8   &92.6   &81.9   &83.2   &67.1   &79.8   &79.0   \\
        % & DiReFT$_{\text{e=6}}$$^*$& 0.031\% &  69.5 & 83.0 & 79.0 & 92.5 & 80.5 & 82.2 & 68.0 & 77.5 & 79.0 \\
        & LoReFT$^\dagger$& 0.031\% &68.3 &83.5& 79.7 &\textbf{92.7} &\textbf{82.6}& 83.2& 67.4& 78.5& 79.5\\
        % & LoReFT$_{\text{e=6}}$$^*$& 0.031\% & 69.3 & 84.4 & 80.3 & \textbf{93.1} & \textbf{84.2} & 83.2 & \textbf{68.2} & 78.9 & 80.2 \\
        \cmidrule{2-12}
        & \textbf{\ourmethodshort{}} & \textbf{0.020\%} & \textbf{69.6}  &\textbf{83.6} &\textbf{80.5} &92.3  &81.1  &\textbf{84.0}  &\textbf{68.1}  &\textbf{80.4} & \textbf{80.0}\\
        \midrule
        & Series$^*$& 1.586\% & 71.8 & 83.0 & 79.2 & 88.1 & 82.4 & 82.5 & 67.3 & 81.8 & 79.5 \\
        & Parallel$^*$& 2.894\% & 72.5 & 84.9 & 79.8 & 92.1 & {84.7} & 84.2 & 71.2 & 82.4 & 81.5 \\
        & LoRA$^*$& 0.670\% & 72.1 & 83.5 & 80.5 & 90.5 & 83.7 & 82.8 & 68.3 & 82.4 & 80.5 \\
        & DoRA$_{\text{half}}$$^\star$ & 0.347\% & 72.5 & 85.3 & 79.9 & 90.1 & 82.9 & 82.7 & 69.7 & 83.6 & 80.8 \\
        LLaMA& DoRA$^\star$& 0.681\% & 72.4 & 84.9 & {81.5} & 92.4 & 84.2 & 84.2 & 69.6 & 82.8 & 81.5\\
        (13B)& SMT$^\ddagger$& 0.680\% & 71.1& 84.4& 81.7 &93.7  &83.2   &86.7  &73.7   &85.2   &82.4  \\
        \cmidrule{2-12}
        & \textbf{\ourmethodshort{}} & \textbf{0.327\%} &\textbf{73.3}&\textbf{87.9}&\textbf{82.7}&\textbf{96.0}&\textbf{86.9}&\textbf{90.2}&\textbf{77.1}&\textbf{88.6}&\colorbox{cyan!30}{\textbf{85.3}} \\
        \cmidrule{2-12}
        & PrefT$^*$& 0.031\% & 65.3 & 75.4 & 72.1 & 55.2 & 68.6 & 79.5 & 62.9 & 68.0 & 68.4 \\
        & DiReFT$^\textbf{+}$& 0.025\% & 70.2 &\textbf{86.6}  &\textbf{82.5} &\textbf{95.0}  &85.2  &86.3  &73.5  &84.4  &83.0  \\
        % & DiReFT$_{\text{e=6}}$$^*$& 0.025\% &  71.3 & 86.1 & 80.8 & \textbf{94.6} & 83.6 & 85.5 & 72.9 & 82.7 & 82.2 \\
        & LoReFT$^\dagger$& 0.025\% & 72.0 &85.6& 82.1& 94.8& \textbf{85.3}& 86.9 &73.0& 85.0& 83.1 \\
        % & LoReFT$_{\text{e=6}}$$^*$& 0.025\% & 72.1 & {86.3} & {81.8} & {95.1} & {87.2} & {86.2} & {73.7} & {84.2} & {83.3} \\
        \cmidrule{2-12}
        & \textbf{\ourmethodshort{}} & \textbf{0.016\%} &\textbf{73.0}   &86.4 &82.2 & 94.5 & 84.0 &\textbf{87.6}  &\textbf{74.5}  &\textbf{86.0} &\textbf{83.5} \\
        \midrule
        & LoRA$^*$& 0.826\% &  69.8 & 79.9 & 79.5 & 83.6 & 82.6 & 79.8 & 64.7 & 81.0 & 77.6 \\
        & DoRA$_{\text{half}}$$^\star$ & 0.427\% & {72.0} & 83.1 & 79.9 & 89.1 & 83.0 & 84.5 & 71.0 & 81.2 & 80.5 \\
        & DoRA$^\star$& 0.838\% &  71.8 & 83.7 & 76.0 & 89.1 & 82.6 & 83.7 & 68.2 & {82.4} & 79.7 \\ 
        Llama2& SMT$^\ddagger$& 0.840\% & 72.0& 83.8& 80.8 &93.3  &82.8   &86.7  &74.0   &81.0   &81.8  \\
        \cmidrule{2-12}
        (7B)& \textbf{\ourmethodshort{}} & \textbf{0.404\%} &\textbf{73.6}   &\textbf{86.5} &\textbf{81.1} & \textbf{94.8} &\textbf{87.8}  &\textbf{89.1}  &\textbf{75.9}  &\textbf{85.6} &\colorbox{cyan!30}{\textbf{84.3}} \\
        \cmidrule{2-12}
        & DiReFT$^\textbf{+}$& 0.031\% &68.2   &\textbf{83.4}  &\textbf{79.8}   &\textbf{93.4}   &83.1   & 84.6  &\textbf{70.3}   &79.4 &80.3  \\
        % & DiReFT$_{\text{e=6}}$$^*$& 0.031\% &  70.8 & 83.6 & 80.2 & 93.6 & 82.1 & 84.8 & 70.4 & 81.5 & 80.9 \\
        & LoReFT$^\textbf{+}$& 0.031\% &  66.6 &81.8  &79.3   &\textbf{93.4}   &82.6   &83.0   &70.2   &80.8 & 79.7\\
        % & LoReFT$_{\text{e=6}}$$^*$& 0.031\% & 71.1 & 83.8 & {80.8} & {94.3} & {84.5} & {85.6} & {72.2} & 82.3 & {81.8} \\
        \cmidrule{2-12}
        & \textbf{\ourmethodshort{}} & \textbf{0.020\%} &  \textbf{71.4} &82.8 &\textbf{79.8} &93.3  &\textbf{84.0}  &\textbf{85.4}  &70.1  &\textbf{81.2} &\textbf{81.0}\\
        \midrule
                & LoRA$^*$& 0.700\% & 70.8 & 85.2 & 79.9 & 91.7 & 84.3 & 84.2 & 71.2 & 79.0 & 80.8 \\
        & DoRA$_{\text{half}}$$^\star$ & 0.361\% & 74.5 & 88.8 & 80.3 & 95.5 & 84.7 & 90.1 & 79.1 & 87.2 & 85.0 \\
        & DoRA$^\star$& 0.710\% & 74.6 & \textbf{89.3} & 79.9 & 95.5 & 85.6 & 90.5 & 80.4 & 85.8 & 85.2 \\ 
        Llama3& SMT$^\ddagger$& 0.710\% & 75.7  &88.4 &81.4  &96.2  &88.2   &92.7  &83.2   &88.6   &86.8  \\
        \cmidrule{2-12}
        (8B)& \textbf{\ourmethodshort{}} & \textbf{0.343\%} & \textbf{75.0}  &\textbf{89.3} &\textbf{83.0} &\textbf{96.5}  &\textbf{89.2}  &\textbf{93.0}  &\textbf{82.9}  &\textbf{89.6} &\colorbox{cyan!30}{\textbf{87.3}} \\
        \cmidrule{2-12}
        % & DiReFT$_{\text{e=6}}$$^*$ & 0.026\% &  73.4 & 88.7 & 81.0 & 95.6 & 85.5 & 91.8 & 81.8 & 85.4 & 85.4 \\
        & DiReFT$^\textbf{+}$ & 0.026\% &  73.0  &89.8   &81.4   &96.1   &87.8   &92.3  &79.9   &85.4   &85.7  \\
        % & LoReFT$_{\text{e=6}}$$^*$& 0.026\% & 75.1 & 90.2 & 82.0 & 96.3 & 87.4 & 92.4 & 81.6 & 87.5 & 86.6 \\
        & LoReFT$^\textbf{+}$& 0.026\% &  72.9& 89.1 &81.7 & 96.1 &88.0   &92.0   &80.1   &85.0   &  85.6 \\
        \cmidrule{2-12}
        & \textbf{\ourmethodshort{}} & \textbf{0.017\%} &71.7   &84.9 &81.4 &93.9  &85.4  &88.8  &77.0  &83.8 &83.4 \\
        \bottomrule
    \end{tabular}
    }
    % \vspace{18pt}
\end{table*}

\section{Experiments}
We evaluate \ourmethodshort{} on 23+ datasets spanning commonsense reasoning (\Cref{sec:commonsense}), arithmetic reasoning (\Cref{sec:Arithmetic}), and natural language understanding (\Cref{sec:Natural language understanding}). Experiments cover both encoder-only (RoBERTa-base~\citep{liu2019roberta}) and decoder-only (LLaMA~\citep{touvron2023llama, touvron2023llama2}) models up to 13B parameters. We benchmark against strong PEFT baselines, including BitFit~\citep{ben-zaken-etal-2022-bitfit}, prefix-tuning~\citep{li-liang-2021-prefix}, adapters~\citep{he2021towards}, LoRA~\citep{lora}, DoRA~\citep{liu2024dora}, SMT~\citep{he_sparse_2024}, RED~\citep{wu2024advancing}, DiReFT, and LoReFT~\citep{wu_reft_2024}. Following LoReFT, all models use \texttt{torch.bfloat16} and run on a single NVIDIA A100 or H100 GPU.

Our comparison considers not only benchmark performance but also parameter efficiency. To demonstrate the robustness of our method under varying parameter budget constraints, we categorize all baseline methods into two groups based on the proportion of trainable parameters: those with $\ge 0.1\%$ and those with $< 0.1\%$. Note that these two groups differ by orders of magnitude in the number of trainable parameters, allowing us to assess performance across both relatively high and extremely low parameter budgets. We then compare our \ourmethodshort{} against these baselines under comparable levels of parameter efficiency to ensure a fair and meaningful evaluation. Specifically, we select the top-1 and top-20 input connections per neuron for matching the budget of the two groups, resulting in 0.016\% and 0.327\% trainable parameters on LLaMA-13B, respectively.

\begin{table*}[!ht]
    \centering
    \caption{
    Performance comparison with existing PEFT methods on seven arithmetic reasoning datasets across four models: LLaMA-7B/13B, LLaMA2-7B, and LLaMA3-8B. 
    $^*$Most baseline results are taken from \citet{hu_llm-adapters_2023}. $^\dagger$Results are from \citet{wu_reft_2024}, $^\ddagger$ results are taken from \citet{he_sparse_2024} and $^\star$results are from \citet{liu2024dora}, as they share the same experimental setting with \citet{hu_llm-adapters_2023}. For a fair comparison, our \ourmethodshort{} is also trained for 3 epochs to align with these baselines.
    $^\textbf{+}$When 3-epoch results are not available in the original paper, we re-trained the baselines using the official code and their reported best hyperparameters. All results for our method are averaged over three runs with different random seeds. Our method selects the top-20 and top-1 input connections per neuron for high-budget and low-budget parameter groups, respectively. 
    }
    \label{tab:math_result}
    \resizebox{1\linewidth}{!}{
    \begin{tabular}{clcccccccccl}
        \toprule
        \hline 
        \rowcolor{cyan!30}
        \textbf{Model} & \textbf{PEFT} & \textbf{Params (\%)} & \multicolumn{9}{c}{\textbf{Accuracy} ($\uparrow$)} \\
        % \multirow{2}{*}{\textbf{Model}} & \multirow{2}{*}{\textbf{PEFT}} & \multirow{2}{*}{\textbf{Params}} & \multicolumn{8}{c}{\textbf{Accuracy} ($\uparrow$)} \\
        \midrule \rowcolor{yellow!30}
        \multicolumn{12}{c}{\textbf{\hspace{8cm} Arithmetic Reasoning}} \\
       \cline{4-12}  \rowcolor{yellow!30}
        & & & &\textbf{MulAri} &\textbf{GSM8K} &\textbf{AddSub} &\textbf{AQuA} &\textbf{SinEq}& \textbf{SVAMP}& \textbf{MAWPS}& \textbf{Avg.} \\
        \midrule
       GPT-3.5\textsubscript{175B}$^*$ &  $\quad-$& $-$ & &83.8 &56.4 &85.3  &38.9  &88.1 &69.9  &$-$&70.4   \\
        \midrule
        & Series$^*$ & 1.953\% &&92.8 &33.3& 80.0 &15.0 &83.5& {52.3}&$-$& 59.5 \\
        & Parallel$^*$ & 3.542\% &&94.5 &35.3& 86.6& 18.1 &86.0& 49.6&$-$&61.7  \\
        & LoRA$^*$ & 0.826\% &&95.0 &\textbf{37.5}& 83.3 &18.9& 84.4& 52.1&$-$&61.9  \\
        & SMT$^\ddagger$ & 0.860\% & &91.5& 34.2& 85.8 &23.6 &84.6  &53.6  &$-$&62.2   \\
        & SMT$^\ddagger$ & 1.260\% & &93.4& 35.6& 86.8 &24.2 &85.3  & 54.8 &$-$&63.4   \\
        \cmidrule{2-12}
        LLaMA& \textbf{\ourmethodshort{}} & \textbf{0.404\%} & &\textbf{96.0}& 36.5&\textbf{92.4}  &\textbf{22.0} &\textbf{94.1}  &\textbf{53.2}  &$-$  &\colorbox{cyan!30}{\textbf{65.7}}\\
        \cmidrule{2-12}
        (7B)& PrefT$^*$ & 0.039\%& &$-$ &24.4& $-$& 14.2& $-$& 38.1&63.4&35.0  \\
        & DiReFT$^+$ & 0.031\% && $-$ & 20.5  & $-$ &21.3  & $-$ &39.9 & 68.1& 37.5 \\
        & LoReFT$^\dagger$  & 0.031\% & &$-$ & 21.6& $-$ & {22.4}& $-$ &43.6&69.5&39.3\\ %\cmidrule{2-11}
        \cmidrule{2-12}
        & \textbf{\ourmethodshort{}} & \textbf{0.020\%} & &$-$& \textbf{30.3} &$-$ &  \textbf{22.8}& $-$ & \textbf{48.9} &\textbf{77.7 }&\textbf{44.9}\\
        % & \textbf{\ourmethodshort{}} & \% & $-$&  &$-$ &  & $-$ &  & &\\
        \midrule
        & Series$^*$ & 1.586\% &&93.0& 44.0& 80.5& 22.0& 87.6& 50.8&$-$&63.0 \\
        & Parallel$^*$ & 2.894\%  &&94.3& 43.3 &83.0& 20.5& 89.6& 55.7&$-$&64.4 \\
        & LoRA$^*$ & 0.670\%& &94.8& \textbf{47.5}& 87.3& 18.5& 89.8 &54.6&$-$&65.4  \\ 
        \cmidrule{2-12}
        LLaMA& \textbf{\ourmethodshort{}} & \textbf{0.327\%} &&\textbf{97.5} &43.9 & \textbf{92.2} &\textbf{21.7} &\textbf{93.9} &\textbf{61.4}  &$-$ &\colorbox{cyan!30}{\textbf{68.4}} \\
        \cmidrule{2-12}
        (13B)& PrefT$^*$ & 0.031\% &&$-$& 31.1& $-$ &15.7& $-$& 41.4&66.8&38.8 \\
        & DiReFT$^+$ & 0.025\% && $-$ & 32.1& $-$ &23.2  & $-$ &51.2 &76.1 &46.7  \\
        & LoReFT$^\dagger$  & 0.025\% && $-$ & 35.5 & $-$ & 23.4& $-$ &54.6&81.8&48.8  \\
        \cmidrule{2-12}
        & \textbf{\ourmethodshort{}} & \textbf{0.016\%} && $-$& \textbf{43.0}&$-$  &\textbf{25.6} &$-$  & \textbf{ 56.7}&\textbf{83.6}  &\textbf{52.2}\\
        \midrule
        & \textbf{\ourmethodshort{}} &  \textbf{0.404\% }& &97.8 &39.8 & 91.9&20.5 &96.3  &54.2 &$-$   &\colorbox{cyan!30}{\textbf{66.8}} \\
        \cmidrule{2-12}
        & DiReFT$^+$ & 0.031\% & &$-$ &26.4 & $-$ & \textbf{23.6} & $-$ &48.4 & 71.8&42.6  \\
        LLaMA2& LoReFT$^+$ & 0.031\%& & $-$ &26.2 & $-$ &18.5  & $-$ &46.7 &76.9 & 42.1 \\\cmidrule{2-12}
        (7B)& \textbf{\ourmethodshort{}} & \textbf{0.020}\% && $-$&\textbf{36.1} &$-$  &22.8 &$-$  &\textbf{52.1} & \textbf{82.4}&\textbf{48.4} \\
        \midrule
        & \textbf{\ourmethodshort{}} &  \textbf{0.343\%} & &99.7 &47.8 &92.7 &27.6 &95.7 &60.4 &$-$   &\colorbox{cyan!30}{\textbf{70.7}} \\
        \cmidrule{2-12}
        & DiReFT$^+$ & 0.026\% && $-$ & 57.2& $-$ &\textbf{30.1}  & $-$ &68.6 &87.8 &60.9  \\
        LLaMA3& LoReFT$^+$ & 0.026\% && $-$ & 56.9& $-$ &24.8  & $-$ & 70.9& 88.2& 60.2 \\\cmidrule{2-12}
        (8B)& \textbf{\ourmethodshort{}} & \textbf{0.017\%} && $-$&\textbf{63.7} &$-$  &26.4 &$-$  & \textbf{75.0}& \textbf{88.7} &\textbf{63.5} \\
        \bottomrule
    \end{tabular}
    }
    \vspace{-5pt}
\end{table*}

\subsection{Commonsense reasoning}
\label{sec:commonsense}
Following the experimental protocol of \citet{hu-etal-2023-llm}, we fine-tune LLaMA-7B/13B, LLaMA2-7B, and LLaMA3-8B on \textsc{Commonsense170K}, a composite dataset consisting of eight commonsense reasoning tasks, as described in \Cref{sec:dataset-details}.
Then, 
We evaluate each task on its test set and compare our results with baselines from \citet{hu-etal-2023-llm}, as well as DoRA~\cite{liu2024dora}, DiReFT, LoReFT~\cite{wu_reft_2024}, and SMT~\cite{he_sparse_2024} under the same setting.

\paragraph{Hyperparameter tuning} Inspired by \citet{wu_reft_2024}, we use \textsc{Commonsense15K}, a subset of \textsc{Commonsense170K}, to perform hyperparameter search. The search space is detailed in \Cref{tab:llama_hyperparameters_commonsense15k}.
Specifically, we split \textsc{Commonsense15K} into training and validation sets, as described in \Cref{sec:Preliminary experiments}.
Our hyperparameter search is conducted only on LLaMA-7B, and the best-performing configuration on the validation set is subsequently applied to all other models, including LLaMA-7B/13B, LLaMA2-7B, and LLaMA3-8B, for training on \textsc{Commonsense170K}.

\paragraph{Results}
As shown in \Cref{tab:commonsense_result}, our \ourmethodshort{} achieves state-of-the-art performance under both parameter budget regimes ($\ge 0.1\%$ and $< 0.1\%$).
Notably, under the higher parameter budget setting, \ourmethodshort{} outperforms all baselines by a considerable margin. For example, its average accuracy surpasses the second-best baseline, SMT, by 4\%.
In addition, \ourmethodshort{} remains effective even under the lower parameter budget setting, consistently outperforming other baselines in this regime.

\subsection{Arithmetic reasoning}
\label{sec:Arithmetic}
Following \citet{hu-etal-2023-llm} and \citet{wu_reft_2024}, we fine-tune LLaMA-7B/13B, LLaMA2-7B, and LLaMA3-8B on \textsc{Math10K}, a composite dataset comprising seven arithmetic reasoning tasks, and evaluate each task separately.
The dataset details are provided in Appendix~\ref{sec:arithData}.

\paragraph{Hyperparameter tuning}
Following \citet{wu_reft_2024}, we perform hyperparameter search on the LLaMA-7B model using the \textsc{GSM8K} dataset, which is split into training and validation sets as in \citet{wu_reft_2024}.
The best-performing configuration on the validation set is then applied to all models, including LLaMA-7B/13B, LLaMA2-7B, and LLaMA3-8B—for training on the \textsc{Math10K} dataset.
The full hyperparameter search space is provided in \Cref{tab:llama_hyperparameters_gsm8k} in Appendix.

\paragraph{Results} 
As shown in \Cref{tab:math_result}, \ourmethodshort{} consistently achieves the highest average accuracy across all model sizes and parameter budgets.
Under a higher parameter budget (e.g., 0.327\% on LLaMA-13B), it outperforms all baselines by a clear margin.
For example, \ourmethodshort{} outperforms the second-best baseline, LoRA, by 3\%, while using even fewer trainable parameters.
Even under extremely low budgets (e.g., 0.020\% on LLaMA2-7B), \ourmethodshort{} remains competitive and surpassing other low-budget baselines by up to 6\% with even fewer trainable parameters.

\subsection{Natural language understanding}
\label{sec:Natural language understanding}
We evaluate the effectiveness of our method on the GLUE benchmark~\citep{wang-etal-2018-glue}, a widely used suite of sequence classification tasks for evaluating natural language understanding (NLU), using the RoBERTa-base model.
To ensure fair comparison, we follow the training, evaluation, and hyperparameter tuning procedures in \citet{wu_reft_2024}.

\paragraph{Results} We report the result in  \cref{tab:glue_result_with_std} in the Appendix due to the space limitation. 
It shows \ourmethodshort{} achieves the highest average score across both moderate (0.2674\%) and extreme low-budget (0.0297\%) regimes. Compared to LoRA (0.239\%), it improves the average GLUE score by {+0.7}. Under the extreme budget, it surpasses LoReFT by {+0.8}, RED by {+0.7}, and DiReFT by {+1.8}, despite using fewer parameters. 
Notably, \ourmethodshort{} achieves the best score on 6 out of 8 tasks in the low-budget setting, demonstrating its strong generalization even with minimal adaptation capacity.

%% file: 5_conclusion.tex
\section{Conclusion}

This paper introduced \ourmethodshort{}, a featherlight and scalable fine-tuning framework that activates each neuron's potential through top-$k$ magnitude-based weight selection. 
By inheriting the performance benefits of sparse tuning and the memory efficiency of addition-based methods, \ourmethodshort{} avoids structural modifications, runtime masking, and full-gradient computation. 
Its static selection of high-magnitude weights per neuron enables task-agnostic, fine-grained adaptation with significantly reduced memory and computational overhead. 
Empirical results across diverse reasoning and language understanding tasks show that \ourmethodshort{} surpasses strong adaptation baselines, achieving robust generalization under an extremely few number of trainable parameters and tight memory budgets.

%% file: 7_limitation.tex
\section{Limitations}
\label{sec:limitations}

While \ourmethodshort{} demonstrates strong empirical performance across diverse tasks and architectures, our current evaluation is limited to models up to 13 billion parameters. 
We anticipate that the benefits of our method may further amplify at larger scales, but assessing its efficacy on models beyond 13B remains an important direction for future work. Evaluating scalability and stability under extreme model sizes is critical for deployment in real-world, high-capacity systems. 
Also, we expect future research to verify \ourmethodshort{} on VLM models with vision-related tasks.

%% file: 6_appendix.tex
\section{Datasets}\label{sec:dataset-details}
Our experiments primarily target both natural language understanding (NLU) and natural language generation (NLG) tasks. Specifically, we evaluate commonsense reasoningand arithmetic reasoning for NLG task and GLUE for NLU task.

\subsection{Commonsense Reasoning}
\label{sec:commonsense-datasets}
Commonsense reasoning is an important task in multimodal learning \cite{zhang-etal-2023-ck}. We evaluate the our method on eight widely-used datasets that span various forms of open-ended question answering:
\begin{itemize}
\item \textbf{BoolQ}\cite{clark-etal-2019-boolq}: a yes/no question-answering dataset composed of naturally occurring questions. Following prior work, we remove the associated passages to focus solely on the question context.
\item \textbf{PIQA}\cite{bisk2020piqa}: designed to test physical commonsense, this dataset requires selecting the most plausible action in a given hypothetical situation.
\item \textbf{SIQA}\cite{sap-etal-2019-social}: targets social commonsense by asking models to reason about human actions and their social consequences.
\item \textbf{HellaSwag}\cite{zellers2019hellaswag}: involves selecting the most coherent sentence completion given a narrative context, emphasizing grounded commonsense inference.
\item \textbf{WinoGrande}\cite{sakaguchi2021winogrande}, inspired by the Winograd Schema Challenge\cite{levesque2012winograd}, tests pronoun resolution in context, requiring fine-grained commonsense reasoning.
\item \textbf{ARC-Easy (ARC-e)}\cite{clark2018think}: a benchmark of multiple-choice elementary science questions with relatively straightforward reasoning.
\item \textbf{ARC-Challenge (ARC-c)}\cite{clark2018think}: a more difficult subset of ARC designed to be resistant to simple co-occurrence-based solutions.
\item \textbf{OpenBookQA (OBQA)}~\cite{mihaylov2018can}: a knowledge-intensive QA dataset requiring multi-hop reasoning across both textual context and external knowledge.
\end{itemize}

We adopt the same experimental protocol as described in \citet{hu-etal-2023-llm}, aggregating the training sets of the above datasets into a unified corpus referred to as \textsc{Commonsense170K}. Fine-tuning is conducted on this joint dataset, and evaluation is performed on the individual test sets of each benchmark. Detailed dataset statistics and simplified training examples are also available in \citet{hu-etal-2023-llm}.

\subsection{Arithmetic reasoning}
\label{sec:arithmetic-datasets}
We train and evaluate our method using seven benchmark datasets that span a diverse range of mathematical word problem domains:

\begin{itemize}
\item \textbf{AddSub}\cite{hosseini-etal-2014-learning}, a dataset composed of elementary arithmetic problems involving addition and subtraction.
\item \textbf{AQuA}\cite{ling2017program}, which presents algebraic word problems in a multiple-choice format.
\item \textbf{GSM8K}\cite{cobbe2021training}, consisting of grade-school math problems that require multi-step reasoning.
\item \textbf{MAWPS}\cite{koncel-kedziorski-etal-2016-mawps}, a collection of math word problems with varied linguistic and arithmetic complexity.
\item \textbf{MultiArith}\cite{roy-roth-2015-solving}, featuring problems that demand reasoning through multiple arithmetic steps.
\item \textbf{SingleEq}\cite{koncel-kedziorski-etal-2015-parsing}, which includes math problems that can be solved by formulating a single equation of varying complexity.
\item \textbf{SVAMP}~\cite{patel-etal-2021-nlp}, designed to test a model’s robustness to structural variations in math problem formulations by rephrasing original problems in a challenging way.
\item \textbf{MAWPS}~\cite{koncel-kedziorski-etal-2016-mawps} is a collection of math word problems of varying complexity, involving basic arithmetic operations such as addition, subtraction, multiplication, and division. Each instance is annotated with both the natural language problem and its corresponding symbolic equation, facilitating studies in semantic parsing and numerical reasoning.
\end{itemize}

We follow the experimental design of \citet{hu-etal-2023-llm}, which also provides dataset statistics and representative examples for each benchmark. Our training is conducted on a unified dataset—\textsc{Math10K}—which comprises training samples from four datasets: GSM8K, MAWPS, MAWPS-single, and AQuA. Following \citep{wu_reft_2024}, we compare our method with LoReFT and DiReFT on four tasks: GSM8K, MAWPS, SVAMP and AQuA and with other baselines on all above tasks except for MAWPS \citet{hu-etal-2023-llm}.

\subsection{Natural language understanding}
\label{sec:glue-datasets}
Following the evaluation protocol established by \citet{wu2024advancing}, we ensure a fair assessment on the GLUE validation set by partitioning it into two subsets. One subset, determined using a fixed random seed, is reserved for in-training evaluation, while the other is used exclusively for final testing. After each training epoch, we evaluate the model on the in-training subset and select the checkpoint with the best performance across all epochs for testing. For datasets with relatively large validation sets (i.e., QQP, MNLI, and QNLI), we randomly sample 1,000 instances for in-training evaluation. For smaller datasets, we use 50\% of the validation data for this purpose. As for the evaluation metrics, we adopt the Matthews correlation coefficient for CoLA, the Pearson correlation coefficient for STS-B, and classification accuracy for the remaining tasks. For MNLI, we report results on the matched subset only.

\section{Results on natural language understanding tasks}
\input{table/results-glue-with-std}

we evaluate our \ourmethodshort{} on the GLUE for natural language understanding tasks. The results are shown in \cref{tab:glue_result_with_std}.  

% \section{Results of commonsense and math reasoning on LLaMA3-8B}
% \input{table/commonsenseAndmath_llama3}
% The results of commonsense reasoning and math reasoning are shown in \Cref{tab:commonsenseAndmath_llama3}

\section{Hyperparameters}

\subsection{Hyperparameter tuning and decoding strategy}\label{sec:hparam}

\input{table/hypersearch-gsm8k}

\paragraph{Arithmeric reasoning}

Following \citet{wu_reft_2024}, we adopt their training and validation splits of the \textsc{GSM8K} dataset.
Models are trained on the training set, and hyperparameters are selected based on performance on the validation set.
All hyperparameters are tuned using LLaMA-7B, and the resulting configuration is directly applied to LLaMA-13B, LLaMA2-7B, and LLaMA3-8B without additional tuning.
We use a maximum sequence length of 512 tokens during training and hyperparameter tuning, and generate up to 32 new tokens during inference.
\Cref{tab:llama_hyperparameters_gsm8k} summarizes the full hyperparameter search space.

\paragraph{Dataset}
\label{sec:arithData}
\textsc{Math10K} is annotated with language model-generated chain-of-thought reasoning steps. In the tasks, models are required to generate a chain of thought~\citep{wei2022chain} before producing the final answer. The included tasks are AddSub~\cite{hosseini-etal-2014-learning}, AQuA~\cite{ling2017program}, GSM8K~\cite{cobbe2021training}, MAWPS~\cite{koncel-kedziorski-etal-2016-mawps}, MultiArith~\cite{roy-roth-2015-solving}, SingleEq~\cite{koncel-kedziorski-etal-2015-parsing}, and SVAMP~\cite{patel-etal-2021-nlp}. See \Cref{sec:arithmetic-datasets} for detailed descriptions of each task. For fair comparison, we follow both evaluation settings used in prior work. Specifically, \citet{wu_reft_2024} report results on GSM8K, AQuA, SVAMP, and MAWPS, while \citet{hu-etal-2023-llm} evaluate on MultiArith, GSM8K, AddSub, AQuA, SingleEq, and SVAMP. We compare our method with theirs under both settings to ensure a comprehensive and fair evaluation.

\input{table/hypersearch-commonsense_15k}

\paragraph{Commonsense reasoning} 
Motivated by \citet{wu_reft_2024}, we perform hyperparameter tuning on \textsc{Commonsense15K}, a subset of the full \textsc{Commonsense170K} benchmark. Details of the search space are provided in \Cref{tab:llama_hyperparameters_commonsense15k}.
We divide \textsc{Commonsense15K} into training and validation splits using a ratio 7:3.
Hyperparameter tuning is conducted exclusively on LLaMA-7B, and the optimal configuration identified on the validation set is reused for all other models, including LLaMA-7B/13B, LLaMA2-7B, and LLaMA3-8B, during training on the full \textsc{Commonsense170K} dataset.

For the commonsense reasoning benchmark, we adopt greedy decoding without sampling, as the task requires multi-token classification.
In contrast, for the arithmetic reasoning benchmark, we follow the decoding setup used by \citet{hu-etal-2023-llm}, employing a higher decoding temperature of 0.3.
This change is made to avoid instability issues caused by near-zero token probabilities, which can lead to decoding errors in the \texttt{HuggingFace} implementation when using beam search.

\paragraph{Dataset} 
\label{sec:commondata}
Our comparative experiments are primarily conducted on LLaMA-7B, using two lightweight reasoning datasets to enable rapid evaluation and comparison: \textsc{Commonsense15k} for commonsense reasoning and \textsc{GSM8K} for arithmetic reasoning. \textsc{Commonsense15k} is a subset of the larger \textsc{Commonsense170k} dataset, originally partitioned by \citet{hu-etal-2023-llm}. (1) \textsc{Commonsense170k} comprises eight commonsense reasoning tasks, including BoolQ~\citep{clark-etal-2019-boolq}, PIQA~\citep{bisk2020piqa}, SIQA~\citep{sap-etal-2019-social}, HellaSwag~\citep{zellers2019hellaswag}, WinoGrande~\citep{sakaguchi2021winogrande}, ARC-e, ARC-c~\citep{clark2018think}, and OBQA~\citep{mihaylov2018can}, as detailed in \Cref{sec:commonsense-datasets}. All examples are presented as multiple-choice questions, where the model is required to directly output the correct option without generating intermediate rationales. We adopt the prompt template from \citet{hu-etal-2023-llm}, with an additional string normalization step (removal of leading and trailing whitespace). We split \textsc{Commonsense15k} into training and validation sets using a 7:3 ratio for our experiments. (2) \textsc{GSM8K}\citep{cobbe2021training} dataset comprises grade-school-level arithmetic word problems that require multi-step reasoning to arrive at the correct answer. In contrast to \textsc{Commonsense15k}, solving \textsc{GSM8K} typically involves generating a chain-of-thought \citep{wei2022chain} prior to producing the final answer. We adopt the same prompt template as used in \citet{hu-etal-2023-llm}.

% \paragraph{Instruction following.} We finetune LLaMA-7B on Alpaca-52K~\citep{alpaca} to select hyperparameters. We select the hyperparameter settings based on model performance evaluated with Alpaca-Eval v1.0~\citep{alpaca_eval}, which calculates the win-rate over \texttt{text-davinci-003} by using \texttt{gpt-4-turbo} as the annotator. We use a maximum sequence length of 768 for training and hyperparameter tuning, and a maximum new token number of 2048 for inference. \Cref{tab:alpaca_hyperparameters} describes our hyperparameter search space.

% During inference, we use the same decoding strategy as in RED~\citep{wu2024advancing} to ensure a fair comparison. Specifically, we use greedy decoding without sampling, and use a maximum repetition n-gram size of 5 with a repetition penalty of 1.1.

\paragraph{Natural language understanding.} 
Following \citet{wu_reft_2024}, we perform hyperparameter tuning for each GLUE task individually using both RoBERTa-base.
Hyperparameters are selected based on performance on a held-out validation set with a fixed random seed of 42.
To obtain the final results, we evaluate the models using four additional unseen seeds: {43, 44, 45, 46}.
We adopt the evaluation protocol of \citet{wu2024advancing}. For QQP with RoBERTa-large, due to observed stochasticity across repeated runs with the same seed, we report the best result from three trials for each seed.
As noted by \citet{wu2024advancing}, the evaluation results on RTE are unstable due to the dataset’s small size. We follow their approach and adjust the set of random seeds accordingly to ensure fair comparison.
Additionally, we replace one or two random seeds for CoLA to improve evaluation stability.

\paragraph{Preliminary Analysis} 
\input{table/hypersearch_differentNum}
We compare the mask-based method—which applies binary masks to zero out the gradients of unselected (frozen) parameters—with the addition-based method, which introduces new trainable parameters to bypass the selected ones, rather than tuning them directly, for the purpose of sparse fine-tuning.
This comparison is conducted across three dimensions: effectiveness, GPU memory usage, and training efficiency.
We further investigate the effectiveness of the proposed addition-based method, \ourmethodshort{}, which is designed to ensure that all neurons in the network retain the potential to update their activation states during fine-tuning.
To this end, we analyze the proportion of neurons actively involved in the tuning process and evaluate various parameter selection strategies for determining which neurons are activated.

To ensure a fair comparison between the addition-based and mask-based approaches, as well as across different parameter selection strategies, we conduct a hyperparameter search over a range of learning rates for each experiment using the training set.
The optimal configuration is selected based on performance on the validation set. This step is essential, as PEFT methods are generally sensitive to the choice of learning rate~\cite{wu_reft_2024}.
The complete hyperparameter search space is provided in \Cref{tab:llama_hyperparameters_different_trainable_num}.

\section{Advantages of \ourmethodshort}
\label{sec:advantages}

\ourmethodshort{} significantly reduces the backward computation costs with four core advantages. 
(1) Compared with current sparse training methods \cite{zhou2026efficient, zhang2024gradient, zhang2024proactive}, it achieves \textbf{mask-free sparsity} by statically selecting top-$k$ weights per neuron and storing only a compact set of BF16 deltas and integer indices. This avoids the memory and compute overhead associated with dynamic binary masks commonly used in gradient-based sparse methods, as quantified in Table~\ref{tab:memory_k1}. 
(2) It requires \textbf{no warm-up or task-specific signal}: the magnitude-based selection operates entirely offline and consistently across tasks, eliminating the need for gradient accumulation or adaptive heuristics. 
(3) It ensures \textbf{neuron-level coverage} by allocating trainable updates to every row of the weight matrix. This guarantees that all neurons participate in learning and avoids the dead filter problem often observed in block- or layer-wise pruning. 
(4) It introduces \textbf{no inference-time overhead}: the sparse update tensor $\boldsymbol{\Delta}$ is merged into the original weights post-training, preserving the model's structure, runtime efficiency, and compatibility with standard inference stacks.

\section{Algorithm}
\label{sec:algorithm}

\ourmethodshort{} follows a three-phase procedure designed for efficiency, simplicity, and compatibility with standard inference infrastructure. As shown in Algorithm~\ref{alg:neuroada}, the process begins with an offline selection phase, where the top-$k$ highest-magnitude weights are identified per neuron based on the pretrained weight matrix. This static selection removes the need for gradient-based importance scoring or dynamic masking during training.

During training, only the selected top-$k$ coordinates per neuron are updated, with all other parameters kept frozen. This enables mask-free, neuron-wise sparse adaptation that significantly reduces gradient computation and optimizer state memory. Crucially, \ourmethodshort{} performs updates directly over a small number of stored deltas, requiring no structural changes or auxiliary layers.

After training, the sparse deltas are merged into the frozen weights via an in-place update, resulting in a model that retains its original structure and supports efficient, standard inference. This design ensures minimal runtime overhead while offering strong adaptation capabilities through fine-grained parameter selection.

\begin{algorithm}[t]
\small
\SetAlgoLined
\KwIn{%
pretrained weight matrix $\boldsymbol{\Phi}\!\in\!\mathbb{R}^{d_\text{out}\times d_\text{in}}$,
top-$k$ budget $k\!\ll\! d_\text{in}$,
training mini–batches $\{(\mathbf{x},\mathbf{y})\}$,
learning-rate $\eta$
}
\KwOut{%
adapted model with merged weights $\boldsymbol{\Phi}\!+\!\boldsymbol{\Delta}$
}

\textbf{Phase 1: Offline Top-$k$ Selection}\;
\ForEach(\tcp*[f]{row $\equiv$ neuron}){$i=1,\dots,d_\text{out}$}{
    % \textit{\\ store $k$ index positions}
    \CommentSty{// store $k$ index positions}
    $\mathcal{I}_i \leftarrow \operatorname{TopK}\bigl(\,|\boldsymbol{\Phi}_{i,:}|,\,k\bigr)$ 
   
}
\textbf{Phase 2: Sparse Training (mask-free)}\;
For each neuron $i$, initialize $\boldsymbol{\Delta}_{i,\mathcal{I}_i} \leftarrow 0$\;

\ForEach{mini-batch $(\mathbf{x},\mathbf{y})$}{
    \CommentSty{// forward pass}\;
    $\mathbf{h}\leftarrow (\boldsymbol{\Phi}+\boldsymbol{\Delta})\,\mathbf{x}$\;
    compute loss $\mathcal{L}(\mathbf{h},\mathbf{y})$\;
    \CommentSty{// backward: update only selected entries}\;
    % \textit{// backward: gradients only on stored positions}\;
    \ForEach{$i=1,\dots,d_\text{out}$}{
        $g_i \leftarrow \nabla_{\boldsymbol{\Delta}_{i,\mathcal{I}_i}}\mathcal{L}$\;
        $\boldsymbol{\Delta}_{i,\mathcal{I}_i}\; \mathrel{-}= \eta\, g_i$
    }
}
\textbf{Phase 3: One-shot Merge and Inference}\;
\ForEach{$i=1,\dots,d_\text{out}$}{
    % add the learned delta directly into the frozen row
    $\boldsymbol{\Phi}_{i,\mathcal{I}_i}\;\leftarrow\;
     \boldsymbol{\Phi}_{i,\mathcal{I}_i} \;+\;
     \boldsymbol{\Delta}_{i,\mathcal{I}_i}$\;
}
Delete $\boldsymbol{\Delta}$ from memory.\;

\CommentSty{\# PyTorch (illustrative)}\\
\CommentSty{y = F.linear(x, merged\_weight, bias)}

\caption{\ourmethodshort{}: Sparse top-$k$ neuron-wise adaptation with merge-compatible updates.}
\label{alg:neuroada}
\end{algorithm}

\clearpage

%% file: table/results-glue-with-std.tex
\begin{table*}[!ht]
    \centering
    \caption{Performance comparison with existing PEFT methods on RoBERTa-base for the GLUE benchmark. $^*$Most baseline results are taken from \citet{wu2024advancing}. The result is presented as the mean with standard deviation (SD) over five runs with different random seeds. $^\dagger$Results are from \citet{wu_reft_2024} as it shares the same experimental setting with \citet{wu2024advancing}. For a fair comparison, our \ourmethodshort{} also follows the same setting.
    }
    \adjustbox{max width=\textwidth}{
    \begin{tabular}{clrrrrrrrrrr}
        \toprule
        \multirow{2}{*}{\textbf{Model}} & \multirow{2}{*}{\textbf{PEFT}} & \multirow{2}{*}{\textbf{Params} (\%)} & \multicolumn{8}{c}{\textbf{Accuracy} ($\uparrow$) (\textbf{SD})} \\
        \cmidrule{4-12}
        & & & \textbf{MNLI} & \textbf{SST-2} & \textbf{MRPC} & \textbf{CoLA} & \textbf{QNLI} & \textbf{QQP} & \textbf{RTE} & \textbf{STS-B} & \textbf{Avg.} \\
        \midrule
        & FT &  100\% &  87.3$_{(0.34)}$ & 94.4$_{(0.96)}$ & 87.9$_{(0.91)}$ & 62.4$_{(3.29)}$ & 92.5$_{(0.22)}$ & 91.7$_{(0.19)}$ & 78.3$_{(3.20)}$ & 90.6$_{(0.59)}$ & 85.6 \\ \cmidrule{2-12}
        & Adapter$^*$ & 0.318\% & 87.0$_{(0.28)}$ & 93.3$_{(0.40)}$ & 88.4$_{(1.54)}$ & \textbf{60.9}$_{(3.09)}$ & 92.5$_{(0.02)}$ & \textbf{90.5}$_{(0.08)}$ & 76.5$_{(2.26)}$ & {90.5}$_{(0.35)}$ & {85.0} \\
        & LoRA$^*$ & 0.239\% & 86.6$_{(0.23)}$ & 93.9$_{(0.49)}$ & 88.7$_{(0.76)}$ & 59.7$_{(4.36)}$ & \textbf{92.6}$_{(0.10)}$ & 90.4$_{(0.08)}$ & 75.3$_{(2.79)}$ & 90.3$_{(0.54)}$ & 84.7 \\
        & Adapter\textsuperscript{FNN}$^*$ & 0.239\% & {87.1}$_{(0.10)}$ & 93.0$_{(0.05)}$ & 88.8$_{(1.38)}$ & 58.5$_{(1.69)}$ & 92.0$_{(0.28)}$ & 90.2$_{(0.07)}$ & 77.7$_{(1.93)}$ & 90.4$_{(0.31)}$ & 84.7 \\
        \cmidrule{2-12}
        RoBERTa & \textbf{\ourmethodshort{}} & 0.2674\% & \textbf{87.7}$_{(0.12)}$&	\textbf{94.3}$_{(0.10)}$&	\textbf{89.7}$_{(0.13)}$&	56.1$_{(0.09)}$&	92.0$_{(0.18)}$&	90.2$_{(0.23)}$&	\textbf{82.0}$_{(0.08)}$&	\textbf{91.0}$_{(0.23)}$&	\textbf{85.4}\\
        \cmidrule{2-12}
        Base& BitFit$^*$ & 0.080\% & 84.7$_{(0.08)}$ & {94.0}$_{(0.87)}$ & 88.1$_{(1.57)}$ & 54.0$_{(3.07)}$ & 91.0$_{(0.05)}$ & 87.3$_{(0.02)}$ & 69.8$_{(1.51)}$ & 89.5$_{(0.35)}$ & 82.3 \\
        & RED$^*$ & 0.016\% & 83.9$_{(0.14)}$ & 93.9$_{(0.31)}$ & {89.2}$_{(0.98)}$ & \textbf{61.0}$_{(2.96)}$ & 90.7$_{(0.35)}$ & 87.2$_{(0.17)}$ & 78.0$_{(2.06)}$ & 90.4$_{(0.32)}$ & 84.3 \\
        & DiReFT $^\dagger$ &  0.015\% & 82.5$_{(0.22)}$ & 92.6$_{(0.76)}$ & 88.3$_{(1.23)}$ & 58.6$_{(1.99)}$ & 91.3$_{(0.19)}$ & 86.4$_{(0.27)}$ & 76.4$_{(1.48)}$ & 89.3$_{(0.56)}$ & 83.2 \\
        & LoReFT $^\dagger$ &  0.015\% & 83.1$_{(0.26)}$ & 93.4$_{(0.64)}$ & {89.2}$_{(2.62)}$ & 60.4$_{(2.60)}$ & 91.2$_{(0.25)}$ & 87.4$_{(0.23)}$ & {79.0}$_{(2.76)}$ & 90.0$_{(0.29)}$ & 84.2 \\
            \cmidrule{2-12}
         & \textbf{\ourmethodshort{}} & 0.0297\%  & \textbf{85.1}$_{(0.09)}$ &	\textbf{94.3}$_{(0.16)}$ &\textbf{	90.7}$_{(0.07)}$ &	59.8$_{(0.12)}$ &	\textbf{92.1}$_{(0.12)}$ &	\textbf{88.1}$_{(0.24)}$& 	\textbf{79.1}$_{(0.23)}$ 	&\textbf{90.7}$_{(0.10)}$ &	\textbf{85.0} \\
        \bottomrule\\
    \end{tabular}
    }

    \label{tab:glue_result_with_std}
\end{table*}

%% file: table/hypersearch-gsm8k.tex
% 11765036_0_yahma-llama-7b-hf_math_perCell_mag_add_times_num1_batch_size16_lr9e-3_num_epoch3_target_modules_q_proj_k_proj_v_proj_o_proj_gate_proj_up_proj_down_proj_weight_decay0_warmup_ratio0.06_max_seq_length512

% 11765036_0_yahma-llama-7b-hf_math_perCell_mag_add_times_num20_batch_size8_lr3e-3_num_epoch3_target_modules_q_proj_k_proj_v_proj_o_proj_gate_proj_up_proj_down_proj_weight_decay0.1_warmup_ratio0_max_seq_length512

\begin{table*}[t]
\centering
\caption{Hyperparameter search space for the LLaMA-7B model using our \ourmethodshort{} on the validation set of the  \textsc{GSM8K}. The best-performing settings are \underline{underlined} for selecting top-$1$ and \uwave{wavy underline } for selecting top-$20$ parameters per neuron in the model. Greedy decoding (without sampling) is used throughout the hyperparameter tuning process.}
\resizebox{0.9\linewidth}{!}{
\begin{tabular}{cc}
\toprule
\multicolumn{2}{c}{\textbf{Hyperparameters (LLaMA-7B on \textsc{GSM8K})} } \\ \midrule
Optimizer & AdamW  \\ [0.2cm]
LR & \{6$\times$10$^{-4}$, 9$\times$10$^{-4}$, 1$\times$10$^{-3}$, \uwave{3$\times$10$^{-3}$}, 6$\times$10$^{-3}$, \underline{9$\times$10$^{-3}$}, 1$\times$10$^{-2}$, 3$\times$10$^{-2}$ \} \\ [0.2cm]
Weight decay & \{0\} \\ [0.2cm]
LR scheduler & Linear \\ [0.2cm]
Batch size & \{\uwave{8}, \underline{16}, 32\} \\ [0.2cm]
Warmup ratio & \{\uwave{0.00}, \underline{0.06}, 0.10\} \\ [0.2cm]
Epochs & \{3\} \\ [0.2cm]
Top-$k$ & \{\underline{1}, \uwave{20}\} \\
\bottomrule
\end{tabular}}
\label{tab:llama_hyperparameters_gsm8k}
\end{table*}

%% file: table/hypersearch-commonsense_15k.tex
% 11765036_0_yahma-llama-7b-hf_commonsense_perCell_mag_add_times_num1_batch_size8_lr8e-3_num_epoch3_target_modules_q_proj_k_proj_v_proj_o_proj_gate_proj_up_proj_down_proj_weight_decay0_warmup_ratio0.1_max_seq_length512

% 11765036_0_yahma-llama-7b-hf_commonsense_perCell_mag_add_times_num20_batch_size16_lr7e-4_num_epoch3_target_modules_q_proj_k_proj_v_proj_o_proj_gate_proj_up_proj_down_proj_weight_decay0_warmup_ratio0.06_max_seq_length512

\begin{table*}[t]
\centering
\caption{Hyperparameter search space for the LLaMA-7B model using our \ourmethodshort{} on the validation set of the  \textsc{Commonsense15K}. The best-performing settings are \underline{underlined} for selecting top-$1$ and \uwave{wavy underline } for selecting top-$20$ parameters per neuron in the model. Greedy decoding (without sampling) is used throughout the hyperparameter tuning process.}
\resizebox{0.9\linewidth}{!}{
\begin{tabular}{cc}
\toprule
\multicolumn{2}{c}{\textbf{Hyperparameters (LLaMA-7B on \textsc{Commonsense15K})} } \\ \midrule
Optimizer & AdamW  \\ [0.2cm]
LR & \{\uwave{7$\times$10$^{-4}$}, 9$\times$10$^{-4}$,  2$\times$10$^{-3}$,  4$\times$10$^{-3}$, 6$\times$10$^{-3}$, \underline{8$\times$10$^{-3}$},  1$\times$10$^{-2}$, 2$\times$10$^{-2}$ \} \\ [0.2cm]
Weight decay & \{0\} \\ [0.2cm]
LR scheduler & Linear \\ [0.2cm]
Batch size & \{\underline{8}, \uwave{16}, 32\} \\ [0.2cm]
Warmup ratio & \{0.00, \uwave{0.06}, \underline{0.10}\} \\ [0.2cm]
Epochs & \{3\} \\ [0.2cm]
Top-$k$ & \{\underline{1}, \uwave{20} \} \\
\bottomrule
\end{tabular}}
\label{tab:llama_hyperparameters_commonsense15k}
\end{table*}

%% file: table/hypersearch_differentNum.tex
\begin{table*}[t]
\centering
\caption{Hyperparameter search space for the LLaMA-7B model using our \ourmethodshort{} on the validation set of the  \textsc{GSM8K} and \textsc{Commonsense15k} for different number of trainable parameters. Greedy decoding (without sampling) is used throughout the hyperparameter tuning process.}
\resizebox{0.9\linewidth}{!}{
\begin{tabular}{cc}
\toprule
\multicolumn{2}{c}{\textbf{Hyperparameters on LLaMA-7B for different number of trainable parameters} } \\ \midrule
Optimizer & AdamW  \\ [0.2cm]
LR & \{6$\times$10$^{-5}$, 9$\times$10$^{-5}$, 5$\times$10$^{-4}$, 7$\times$10$^{-4}$, 9$\times$10$^{-4}$, 3$\times$10$^{-3}$, 6$\times$10$^{-3}$, 9$\times$10$^{-3}$, 1$\times$10$^{-2}$ \} \\ [0.2cm]
Weight decay & \{0\} \\ [0.2cm]
LR scheduler & Linear \\ [0.2cm]
Batch size & \{16\} \\ [0.2cm]
Warmup ratio & \{0.06\} \\ [0.2cm]
Epochs & \{3\} \\ [0.2cm]
Top-$k$ & \{1 5 10 20 50 100 300 500\} \\
\bottomrule
\end{tabular}}
\label{tab:llama_hyperparameters_different_trainable_num}
\end{table*}